\pdfoutput=1

\documentclass[11pt]{article}

\usepackage[]{acl}

\usepackage{times}
\usepackage{latexsym}
\usepackage{graphicx}

\usepackage[T1]{fontenc}

\usepackage[utf8]{inputenc}

\usepackage{microtype}

\usepackage{CJKutf8}
\title{Seeking Diverse Reasoning Logic: Controlled Equation Expression Generation for Solving Math Word Problems}

\author{Yibin Shen\thanks{\quad This denotes equal contribution.}$\ \,^1$, Qianying Liu\footnotemark[1]$\ \,^{2}$, Zhuoyuan Mao$^2$, Zhen Wan$^2$, Fei Cheng$^2$ and Sadao Kurohashi$^2$\\
$^1$ Meituan\\
$^2$ Graduate School of Informatics, Kyoto University \\
  {\tt
  shenyibin@meituan.com;}
  {\tt
  \{ying,zhuoyuanmao,zhenwan\}@nlp.ist.i.kyoto-u.ac.jp;}\\{\tt \{feicheng, kuro\}@i.kyoto-u.ac.jp} \\
  }

\begin{document}
\maketitle
\begin{abstract}
To solve Math Word Problems, human students leverage diverse reasoning logic that reaches different possible equation solutions. However, the mainstream sequence-to-sequence approach of automatic solvers aims to decode a fixed solution equation supervised by human annotation.
In this paper, we propose a controlled equation generation solver by leveraging a set of control codes to guide the model to consider certain reasoning logic and decode the corresponding equations expressions transformed from the human reference.
The empirical results suggest that our method universally improves the performance on single-unknown (Math23K) and multiple-unknown (DRAW1K, HMWP) benchmarks, with substantial improvements up to 13.2\% accuracy on the challenging multiple-unknown datasets. \footnote{Our code is  available at \url{https://github.com/yiyunya/CTRL-MWP}. } 
\end{abstract}

\section{Introduction}

Solving Math Word Problems (MWPs) is the task of obtaining mathematical solutions from natural language text descriptions. Recent studies leverage sequence-to-sequence (seq2seq) neural networks (NNs) for solving MWPs, which take in the text as the input
and decodes the corresponding human-annotated equation reference, which can further calculate the answer value~\citep{wang-etal-2017-deep}. 
While promising results have been reported for single-unknown variable problems by designing task specialized encoder and decoder architectures~\cite{wang2018mathdqn, wang2019template, xie2019goal, liu-etal-2019-tree, guan-etal-2019-improved, zhang2020graph, ijcai2020-555, shen-jin-2020-solving}, using pre-trained models~\cite{tan2021investigating,liang2021mwpbert} and leveraging auxiliary tasks~\cite{liu2020reverse, shen2021generate, li2022seeking}, various studies for a more challenging setting, MWPs with multiple-unknowns
have recently been developed~\cite{upadhyay-chang-2017-annotating,qin-etal-2020-semantically,cao2021bottom,qin-etal-2021-neural}. 
\begin{figure}[t]
  \centering
  \includegraphics[width=\linewidth]{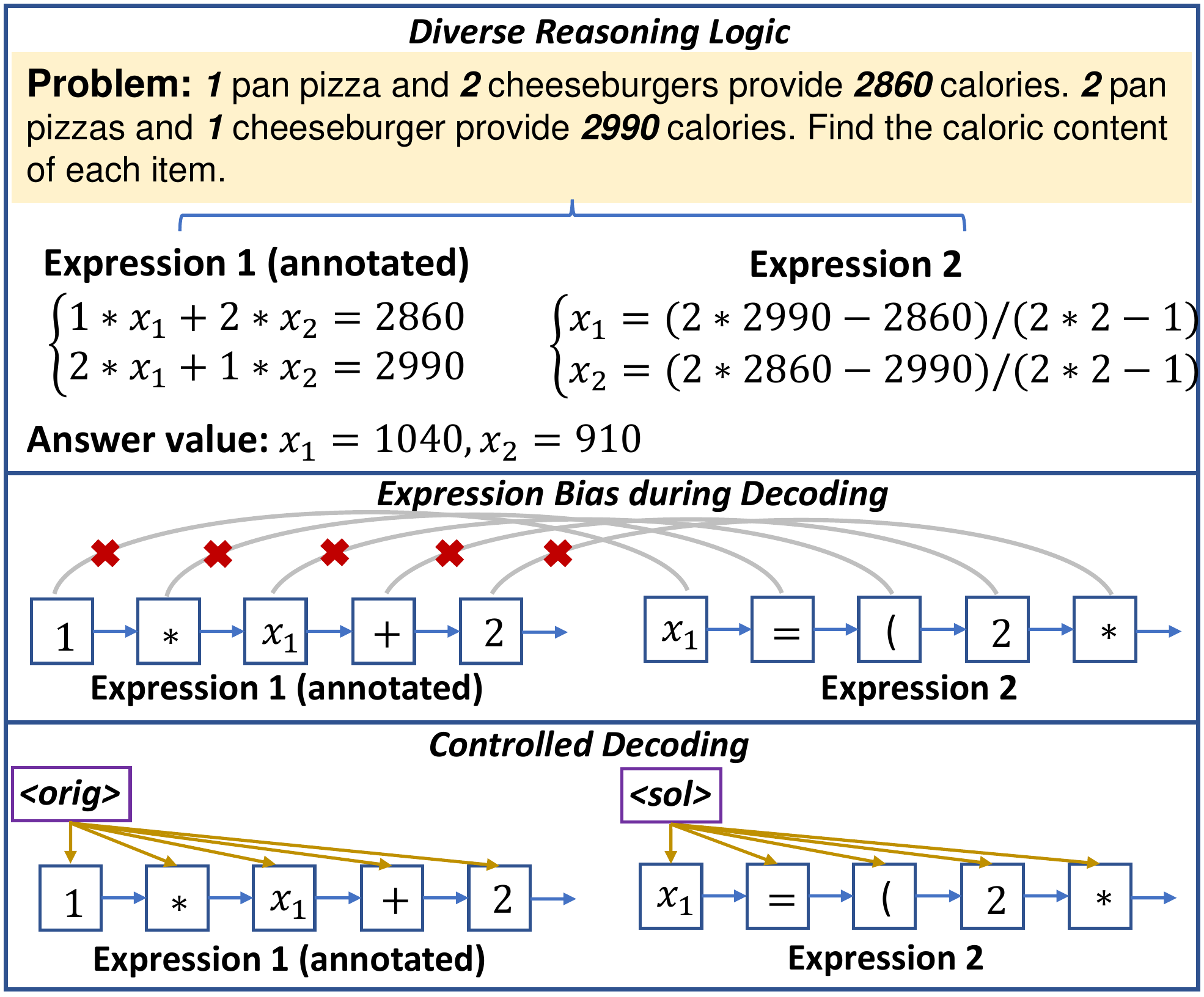}
  \caption{Example of diverse reasoning logic, expression bias, and our controlled expression generation. \textit{<orig>} and \textit{<sol>} are the pre-defined control codes. }
  \label{fig:ex}
  \end{figure}
\textcolor{black}{For human students in practice, they intuitively use diverse reasoning logic to solve MWPs. Students could consider the MWP solution from different aspects by considering diverse equivalence relations in the MWP.}
\textcolor{black}{As we show in the upper of Figure \ref{fig:ex}, we can solve this problem in at least two different reasoning logic: As shown on the left side, the equation set is formed by the first reasoning logic of ``considering the equivalence relation of the two sums of the cheeseburger and pizza calories given in the question''; or as shown in the right side, we can follow a second reasoning logic ``considering first only the equivalence relation of caloric content of the cheeseburger by offsetting the calories from the pizza''.
Such diverse reasoning logic could lead to diverse equation expressions, that the solution equation is written in various mathematically equivalent forms, such as expression 1 and expression 2 in the example.
However, previous studies share a long-lasting limitation that they force the solver to decode a fixed equation expression supervised by human annotation.
}
The fixed equation expression supervision used in previous studies ignores diverse mathematical reasoning, which is especially common for human students in multiple-unknown problems and complex single-unknown problems.


Meanwhile, directly introducing diverse equation expressions to the seq2seq framework in a data augmentation manner could further aggravate the issue of expression bias, which refers to the discrepancy between the annotated equation expression and the model's correct prediction expression. 
As shown in the middle of Figure \ref{fig:ex}, even if the model makes the correct prediction of the problem, the training loss accumulated by diverse expressions could be enormous. \citet{wang-etal-2018-translating} propose an equation normalization that reorders the variables in the equations as close as possible to their order in the input text. While their method could reduce the expression bias issue, they ignore the inherent diverse mathematical reasoning and limits to considering single-unknown problems.

Enlightened by recent methods in controlled text generation, which uses a control code to influence the style and topic of subsequent generated text~\cite{keskar2019ctrl,autoprompt:emnlp20}, 
we propose a new training paradigm, where a control code guides the decoding process to consider one type of mathematical reasoning logic and decode the corresponding equation expression. 
As shown in the bottom Figure \ref{fig:ex}, the \textit{<sol>} control code guides the model to consider 
the direct solution of each individual unknown $x_1$ and $x_2$.
Not only can it reduce the expression bias problem since the control code can provide guidance for the reasoning logic, but also training on the diverse equation expressions guided by the control codes can lead to better interpretation of the MWPs by considering diverse reasoning logic.
We design various control codes for both single-unknown and multiple-unknown settings to allow the model to understand different reasoning orders.
We conduct experiments on a single-unknown benchmark Math23K and two multiple-unknown benchmarks DRAW1K and HMWP. Experimental results show that our method improves the performance of both settings, with a more significant improvement in the challenging multiple-unknown setting.


  \section{Methodology}
  
  For each math word problem holding an original equation set $(e_1,e_2,...)$, we generate new equation expressions based on five types of diverse mathematical reasoning logic considering the ordering logic of given variables $\{n_i\}$ and unknown variables $\{x_j\}$. $i$ and $j$ denote the ordered indices that the variables appear in the text. We then assign a corresponding control code to the equation expressions.
  The MWP solving model takes in the text and control code, and then is trained to predict the corresponding equation expression. 
  
      \begin{figure*}[t]
  \centering
  \includegraphics[width=0.8\linewidth]{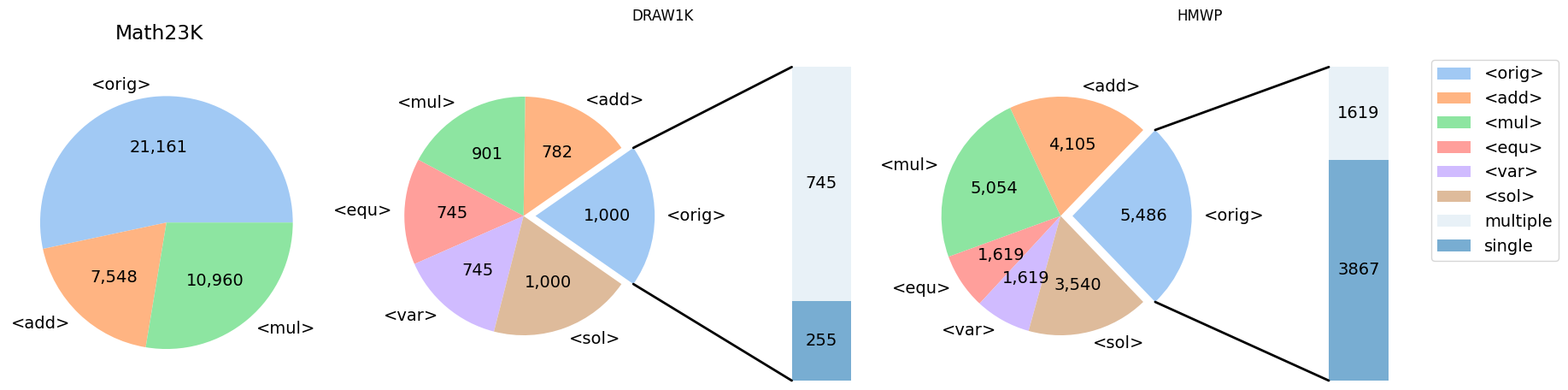}
  \caption{Statistics of datasets and the usage of control codes.}
  \label{fig:stat}
  \end{figure*}
  
  \subsection{Control Codes}
  
We consider the diverse mathematical reasoning logic in two aspects. The first aspect considers diverse reasoning orders of given variables,
which reflects in the diverse expressions of the commutative law and solution form.
For example, $n_1*x_1 = n_2$ could be transformed to the solution form $x_1 = n_2 / n_1$ which does not effect the mathematical equivalency.
This approach is valid for both multi-unknown and single-unknown problems.
The second aspect considers diverse reasoning orders of unknown variables,
which reflects in the diverse expressions of equivalent equation sets. For example, swapping the equation order in the equation set does not affect the mathematical equivalency. This approach is valid for multi-unknown problems.


We preprocess the equation annotations with Sympy~\cite{10.7717/peerj-cs.103} so that they follow a predefined order similar to \citet{wang-etal-2018-translating}. Then we generate different types of equation expressions based on these preprocessed equations.

For the first aspect, we consider three types of diverse equation expressions.

\begin{itemize}
    \item \textbf{Commutative Law of Addition \textit{<add>}} 
    We traverse the equation in prefix order, and swap the left and right subtrees of the addition operators.
    For example,
    $x_1 = n_1 + n_2 + n_3$ would be swapped two times. We first swap the two subtrees $n_1$ and $n_2$ of the first addition operator to $x_1 = n_2 + n_1 + n_3$, and then swap the two subtrees $n_2 + n_1$ and $n_3$ of the second operator to $x_1 = n_3 + n_2 + n_1$.
    \item \textbf{Commutative Law of Multiplication \textit{<mul>}} Similarly, we traverse the equation in prefix order, and swap the left and right subtrees of the multiplication operators. For example,
    from $x_1 = n_1 * n_2 * n_3$ to $x_1 = n_3 * n_2 * n_1$.
     \item \textbf{Solution Form \textit{<sol>}} We consider a mathematical reasoning method that directly considers the solution of each unknown variable. For example,
     from $n_1/x_1 = n_2$ to $x_1 = n_1 / n_2$.
    
\end{itemize}


For the second aspect, we consider two types of diverse equation expressions.

\begin{itemize}
    \item \textbf{Equation Swapping \textit{<equ>}} We swap the multiple-unknown equations in sequential order, which means given a list of equations $(e_1,e_2,...e_n)$, we swap them to the order $(e_n,e_1,e_2,...e_{n-1})$.
    \item \textbf{Unknown Variable Swapping \textit{<var>}} Similarly, we swap the multiple unknown variables in sequential order, which means given a list of unknown variables in the equation $(x_1,x_2,...x_n)$, we change the correspondence between them and the unknown variables in the original question, that the unknown variables in the new equation $(x_1^s,x_2^s,...x_n^s)$ follows $x_1^s$ denotes $x_n$ and $x_i^s$ denotes $x_{i-1}$ for other indices. For example,
    from $n_1 * x_1 + n_2 * x_2 = 0$ to $n_1 * x_2 + n_2 * x_1 = 0$.
    
\end{itemize}
 

To incorporate the control codes for guiding the equation expression decoding, we follow studies in controlled text generation~\cite{keskar2019ctrl} and append a control code to the encoder input. 
We use an independent special token for each expression category as the control code, such as \textit{<add>}, including \textit{<orig>} for the example of the original equation expression.
\textcolor{black}{We use the prediction of the original equation expression control code \textit{<orig>} for test inference since it has the most training examples.}


  \subsection{MWP solving model}

Solving multiple-unknown problems usually requires equation sets, which are challenging to generate. To tackle this problem,  we follow the decoding target paradigm of \citet{qin-etal-2020-semantically}, which introduces a Universal Expression Tree (UET) to represent multiple-unknown equation sets uniformly as an expression tree by using a dummy node as the head of the equation set. UET can also handle single-unknown problems in a unified manner.
  
   For the solver model, we use two strong baseline models for experiments.
   For the first model, we leverage a seq2seq pre-trained language model BART~\cite{lewis-etal-2020-bart,shen2021generate} as the solver model, which has reported promising results for text generation tasks. The encoder takes in the textual input and generates high-quality representations of the problem text. The decoder generates the UET based on these representations.
   
   For the second model, we follow \citet{li2022seeking} and use BERT-GTS as MWP solving model. We leverage the contextual pre-trained language model BERT as the encoder, and use a Goal-driven tree-structured MWP solver (GTS) ~\cite{xie2019goal} based on Long-Short-Term-Memory networks (LSTM) as the decoder.
   

      \begin{table*}
\centering
\begin{tabular}{lcccc}
\hline
\textbf{Model} &Math23K & DRAW &HMWP  \\
\hline
GTS~\cite{xie2019goal}&75.6&39.9&44.6\\
G2T~\cite{zhang2020graph}&77.4&41.0&45.1\\
SAU-Solver~\cite{qin-etal-2020-semantically}&-& 39.2&44.8\\
BART$^\dagger$~\cite{shen2021generate} & 80.4&32.1&41.5\\
BERT-GTS$^\dagger$~\cite{li2022seeking} &82.6&42.2&48.3\\

\hline
Controlled BART & 82.3&45.3&47.9\\
Controlled BERT-GTS &\textbf{84.0}& \textbf{50.2}&\textbf{54.1}\\
\hline
\end{tabular}
\caption{Results on MWP datasets. $\dagger$ denotes our implementation results.}
\label{tab:result}
\end{table*}

      \begin{table}
\centering
\begin{tabular}{lcccc}
\hline
\textbf{Model} &Math23K & DRAW &HMWP  \\
\hline
BERT-GTS&82.6&42.2&48.3\\
+ \textit{<add>} &83.0&46.8&50.8\\
+ \textit{<mul>}&83.3&47.6&51.9\\
+ \textit{<sol>}&-&46.3&50.5\\
+ \textit{<equ>}&-&48.3&50.1\\
+ \textit{<var>}&-&47.4&50.1\\
\hline
All &\textbf{84.0}&\textbf{50.2}&\textbf{54.1}\\
- \textit{code} & 83.3&49.6&49.6\\
\hline
\end{tabular}
\caption{Ablation Study on MWP datasets. \textit{+ <control code>} denotes using only one control code. \textit{All} denotes using all control codes. \textit{- code} denotes using the examples as data augmentation without control codes.}
\label{tab:ab}
\end{table}

  \section{Experiments}
  \subsection{Datasets}

  We evaluate our proposed method on one single-unknown Chinese dataset \textbf{Math23K}~\cite{wang-etal-2017-deep} and two multiple-unknown datasets, \textbf{DRAW1K}~\cite{upadhyay-chang-2017-annotating} in English and \textbf{HMWP}~\cite{qin-etal-2020-semantically} in Chinese.
We show the statistics of overall data size, single and multiple unknown problem size, and the usage of control codes of the datasets in Figure \ref{fig:stat}. The five control code methods are enumerated for each example to generate new equation expressions. While \textit{<sol>} is applicable for both single-unknown and multiple-unknown problems, the annotation schema in Math23K uses the Solution Form, which corresponds to \textit{<orig>}, that no more further equation expressions are generated for\textit{<sol>}. We use from 1.87 to 6.15 times of original data examples size for training on the three datasets.


  \subsection{Results}
  
We show our experimental results on the three datasets in Table \ref{tab:result}. We compare our results with three models: \textbf{GTS} uses an LSTM encoder and decoder, which considers tree structure information during decoding; \textbf{G2T} uses a Graph Neural Network that considers quantity information as the encoder and similar tree decoder; \textbf{SAU-Solver} introduces a semantically-alignment to the target vocabulary of the equations to improve the GTS decoder. As we can see, our method outperforms the baseline for both models on all datasets. The accuracy of different models gains improvement from 1.8\% to 1.9\% for single-unknown problems and from 4.8\% to 13.2\% for multiple-unknown problems. The results demonstrate the effectiveness of our method, especially for multiple-unknown problems. 


  \subsection{Ablation Study}
  
  We conduct further analysis on the more effective model BERT-GTS. In Table \ref{tab:ab}, we show the ablation study using different control codes. As shown in the Table, using each control code individually can improve the model's prediction. \textit{<mul>} is particularly effective for all datasets since it has an extensive example size for each dataset. Using all control codes together further boosts the model performance by providing diverse mathematical reasoning logic as guidance.
  
  We also show the results of removing the control codes and solely using the diverse equation expressions in a data augmentation manner in Table \ref{tab:ab}. Solely introducing diverse mathematical reasoning logic can also improve the model performance compared to the baseline model. However, the expression bias problem limits the performance since training loss could accumulate for diverse equation expressions. By incorporating control codes to guide the decoding process, our method can consider diverse reasoning logic and reduce the expression bias problem in the meantime.
 
 \subsection{Study on Variable Size}
 
\begin{figure}[t]
  \centering  \includegraphics[width=0.7\linewidth]{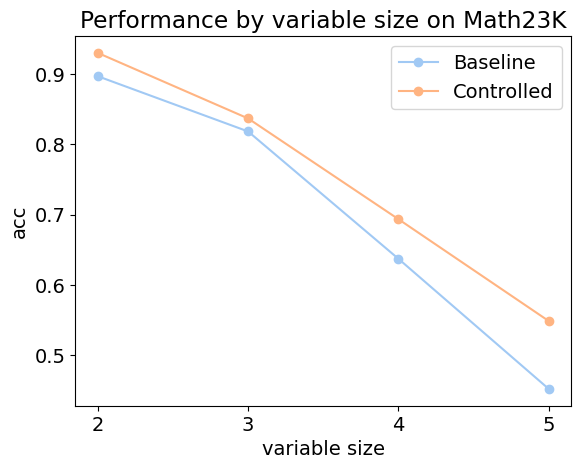}
  \caption{Performance on different given variable sizes.}
  \label{fig:size}
  \end{figure}

 We show the performance on different given variable sizes of the BERT-GTS baseline model and our controlled equation generation method on Math23K in Figure \ref{fig:size}. As the variable size grows, the problem becomes more complex, and the performance gap between our method and the baseline becomes more significant. Our method can incorporate diverse equation expressions to help the model learn mathematical reasoning logic.
 
  \begin{CJK*}{UTF8}{gbsn}
      \begin{table*}
\centering
\begin{tabular}{lccc}
\hline
\textbf{Category} &English &Chinese  \\
\hline
\textit{<add>}&Swap addition operands&加法交换律\\
\textit{<mul>}&Swap multiplication operands&乘法交换律\\
\textit{<sol>}&Solution form&以解形式表达\\
\textit{<equ>}&Swap equation order sequentially&交换方程组算式\\
\textit{<var>}&Swap unknown variables order sequentially&交换未知量\\
\hline
\textit{<orig>}&Original Form&原始形式\\
\hline
\end{tabular}
\caption{Description based control codes used for each category.}
\label{tab:dd}
\end{table*}
\end{CJK*}

      \begin{table}
\centering
\begin{tabular}{lcccc}
\hline
\textbf{Model} &Math23K & DRAW &HMWP  \\
\hline
BERT-GTS&82.6&42.2&48.3\\
+ \textit{token} &\textbf{84.0}&50.2&54.1\\
+ \textit{description}&83.3&\textbf{52.1}&\textbf{55.2}\\
\hline
\end{tabular}
\caption{Study on using different control code strategies. \textit{+token} denotes using special tokens. \textit{+description} denotes using a short description text of the category.}

\label{tab:des}
\end{table}
  \subsection{Study on control code strategies}
  \label{sec:dtcc}

  Various studies have shown that natural language style control codes that serve as a description of the target text could benefit the model performance~\cite{keskar2019ctrl, he2020ctrlsum}. In Table \ref{tab:des}, we show the performance of applying a description text based control code for each expression category, such as \textit{Swap addition operands}. We use the description text \textit{Original input} for the origin equation expression \textit{<orig>} category, and also use it for inference at test stage. The detailed descriptions are shown in 
 Table \ref{tab:dd}. Description text based control codes achieve better performance on multiple-unknown datasets, which have more expression categories. Such control codes could be beneficial as more controlled equation generation strategies are applied, which we leave as future work.
  
  \section{Conclusion and Future Work}
  
  In this paper, we introduce diverse mathematical reasoning logic to the seq2seq MWP solver framework using five control codes to guide the solver to predict the corresponding equation expression in a controlled equation generation manner. The approach allows the solver to benefit from diverse reasoning logic beyond the human-annotated fixed solution equation. Meanwhile, the controlled equation generation training paradigm reduces the expression bias problem caused by diverse equation expressions. Experimental results show the effectiveness of our method, outperforming strong baselines on single-unknown (Math23K) and multiple-unknown (DRAW1K, HMWP) datasets.
  
  \textcolor{black}{There exists other controlled equation generation strategies such as such as adding brackets to merge subtraction terms (e.g. from $n_1-n_2-n_3$ to $n_1-(n_2+n_3)$) or combining current control codes to form a new type of equation expression, which potentially could lead to more than 10 controlled equation generation strategies. In addition, considering the prediction of multiple control codes in addition to \textit{<orig>} could further improve the performance results, for example, applying ensemble learning methods such as major voting, or designing rankers to choose a optimal prediction among the prediction of multiple control codes. These problems could be considered as future work of this study.}

\bibliography{anthology,custom}
\bibliographystyle{acl_natbib}

\appendix
\section{Experimental Details}
\label{sec:appendix}

We evaluate Math23K on the standard train test setting. DRAW1K and HMWP are evaluated by 5-cross validation.

For DRAW1K, we use the bert-base pre-trained encoder. For Math23K and HMWP, we use the pre-trained encoder that could be found here \footnote{https://huggingface.co/yechen/bert-base-chinese}.

For Math23K, the max text length is 256, the max equation decoding length is 45, the batch size is 16 and the epochs number is 50. We use AdamW with a learning rate of 5e-5.

For DRAW1K, the max text length is 256, the max equation decoding length is 32, the batch size is 16 and the epochs number is 50. We use AdamW with a learning rate of 5e-5.

For HMWP, the max text length is 1024, the max equation decoding length is 100, the batch size is 8 and the epochs number is 50. We use AdamW with a learning rate of 5e-5.

Experiments are conducted on NVIDIA 3090 and A100(80G). The runtime for the longest experiments is around 6 hours.

\end{document}